\documentclass[letterpaper, 10 pt, conference]{ieeeconf}  

\IEEEoverridecommandlockouts                              

\usepackage{graphics} 
\usepackage{epsfig} 
\usepackage{mathptmx} 
\usepackage{amsmath} 
\usepackage{amssymb}  

\title{\LARGE \bf
The Power of Combined Modalities in Interactive Robot Learning}

\author{Helen Beierling$^{1}$ and Anna-Lisa Vollmer$^{1}$
\thanks{This work was funded by the Deutsche Forschungsgemeinschaft (DFG, German Research Foundation): TRR 318/1 2021 – 438445824.}
\thanks{$^{1}$The authors are with the Medical School Ostwestfalen-Lippe and the Center for Cognitive Interaction Technology, Bielefeld University, Bielefeld, Germany
        {\tt\small firstname.lastname@uni-bielefeld.de}}%
}

\begin{document}

\maketitle
\thispagestyle{empty}
\pagestyle{empty}

\begin{abstract}
This study contributes to the evolving field of robot learning in interaction with humans, examining the impact of diverse input modalities on learning outcomes. It introduces the concept of “meta-modalities,” which encapsulate additional forms of feedback beyond the traditional preference and scalar feedback mechanisms. Unlike prior research that focused on individual meta-modalities, this work evaluates their combined effect on learning outcomes. Through a study with human participants, we explore user preferences for these modalities and their impact on robot learning performance. Our findings reveal that while individual modalities are perceived differently, their combination significantly improves learning behavior and usability. This research not only provides valuable insights into the optimization of human-robot interactive task learning but also opens new avenues for enhancing the interactive freedom and scaffolding capabilities provided to users in such settings.
\end{abstract}

\section{Introduction}
Artificial intelligence (AI) now permeates various aspects of our lives, exemplified by virtual assistants like ChatGPT and suggestion algorithms on social media platforms.
AI systems often rely on user feedback to learn and improve.
This feedback can take various forms, from traditional star ratings to more passive indicators like watch time.

There is a growing demand for intelligent agents that can simplify everyday tasks, particularly in home environments characterized by diverse individual preferences and tasks.

To address this, users, especially lay users, need to be empowered to teach robots new actions according to their preferences.
Interactive learning with users in the loop, based on active feedback, is a common method for achieving this, where scalar and preference feedback are popular options \cite{kaufmann2023survey}.
\begin{figure}
    \centering
    \includegraphics[width = 0.5 \textwidth]{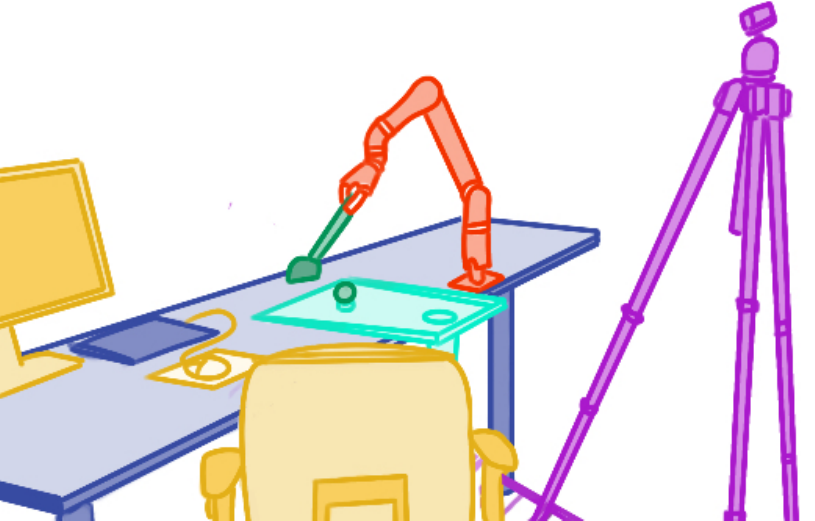}
    \caption{Study setup. Red depicts the used robot platform, green the task relevant objects, purple the camera for study recording, and yellow the objects the user gets in contact with.}
    \label{fig:setup}
\end{figure}
Research suggests that humans are more adept at comparing options than rating them on a scale, making preference feedback a foundational element \cite{hindemith2022interactive}.
Additionally, literature advocates for modalities that allow users to express supplementary feedback to enhance the basis feedback quality and guide actions effectively \cite{thomaz2006reinforcement,liu2023interactive,li2021learning,bajcsy2018learning,celemin2023knowledge,ravichandar2020recent,arzate_cruz_survey_2020}
; we refer to these as "meta-modalities".\\
While the literature presents various beneficial meta-modalities, they have been largely tested in isolation.
Therefore, our study explores whether all modalities are utilized and perceived equally, and whether their combined use offers a learning benefit.

\section{Related Work}

Human inputs in reinforcement scenarios are valuable, given that reward functions customized for specific situations may underperform in diverse contexts \cite{kaufmann2023survey}.
Additionally, it can be challenging to define such functions rationally via reward engineering, especially for tasks based on personal preferences.
There is abundant literature spanning from older approaches~\cite{isbell2001social,knox2008tamer} to more contemporary ones \cite{hindemith2022interactive,ding2023learning} regarding the utilization of human feedback that were able to show positive results with various usage methods of human feedback.
Thus, the approach of incorporating human feedback in reinforcement learning is current and relevant, particularly within the realm of robotics.

Concerning the base modality, we concentrate on the input modalities identified as most common.
In the survey by Kaufmann et al. \cite{kaufmann2023survey}, examining existing approaches for reinforcement learning with direct user feedback, they are classified into preference-based and reward-based learning approaches.
Here, users provide numerical or binary evaluations of the agent's actions, reflecting evaluative feedback. 
In addition, emotions \cite{yu2019interactive,su2023recent}, gestures \cite{su2023recent}, or speech \cite{chen2022real} can also be recognized and used as feedback individually or in combination.
However, this paper focuses on direct evaluative feedback methods.
Our decision to adopt preference as the base feedback method was based on 
a recent study comparing preference and scalar feedback within a reinforcement learning algorithm based on evolutionary black-box optimization in a robot setup \cite{hindemith2022interactive}.
The findings indicated that users were better able to compare two trajectories than provide scalar ratings within their own contextual framework of past ratings for other trajectories.\\
In addition to the base modality, we identified six modalities from the literature that build upon them. 

\textit{Guidance}: An earlier study by Thomaz et al. \cite{thomaz2006reinforcement} highlights the importance of providing users with the option to provide feedback not only for present actions but also for future actions, giving guidance to the robot.
This method not only prevents users from misusing feedback options for guidance but also enhances learning outcomes.

\textit{Correction}: Liu et al. \cite{liu2023interactive} suggested the opposite of guidance, namely correction.
The literature suggests that using corrections for human-in-the-loop robot learning benefits learning \cite{liu2023interactive, li2021learning, bajcsy2018learning, celemin2023knowledge}.

\textit{Demonstration}: Another modality is the possibility to demonstrate an action to the robot to assist or lead the learning process. 
Approaches like kinesthetic teaching are increasingly popular \cite{ravichandar2020recent}.

\textit{Exploration}: 
While the exploration versus exploitation dilemma is a longstanding challenge in machine learning, it becomes particularly pronounced in scenarios where the algorithm receives sparse rewards. 
In such cases, random exploration risks wasting valuable rewards.
However, this challenge can be mitigated by leveraging demonstrations to guide exploration. 
Besides demonstration which we already covered, one can directly incorporate the user's guidance for exploration \cite{arakawa2018dqn,arzate_cruz_survey_2020}.

\textit{Speed}: A modality closely aligned with the action advice concept outlined by Arzate Cruz et al. \cite{arzate_cruz_survey_2020} involves offering guidance on specific attributes of an action.
We selected execution speed as a simple attribute for user guidance.

\textit{Fallback}: Lastly, there is the option to revert to a previous state of action and resume learning from there.
This modality, grounded in the concept of correction or negative guidance, as outlined by Thomaz et al. \cite{thomaz2006reinforcement}.
In instances where the agent
converges to undesired behavior, a fallback modality allows for retracing the explored path and restarting from the best action achieved thus far.

While there is considerable research on individual meta-modalities, their combination remains an open question. A survey focusing on interactive robotics and human-robot interactions, examining trendsetting human feedback methods highlights the open challenge of combining feedback types and their application to reinforcement learning algorithms \cite{arzate_cruz_survey_2020}. 
Similarly, the potential of combining feedback methods in interactive reinforcement has been emphasized \cite{li_human-centered_2019}. 

Our work aims to address this research gap and study the combination of multiple feedback modalities for robot learning in interaction with human users.
Our objective is to investigate whether learning behavior improves, which modalities users utilize, and which modality they prefer.


\section{Methodology}

\subsection{Implementation}
The modalities were implemented on a Kinova Gen2/Jaco (Jaco2) robotic arm \cite{kinova} using ROS Noetic \cite{noetic}.

To represent movements, we utilized an implementation of probabilistic movement primitives (ProMP) \cite{paraschos2013probabilistic, proMP}.
ProMP depict movements based on parameterized distributions, rendering them highly adaptable as they encapsulate entire distributions rather than singular trajectories.
Moreover, they boast a compact design, necessitating adjustments solely to their parameters to accommodate changes in trajectories.
We implemented the PIBB black-box algorithm for learning purposes \cite{stulp2012policy}.
The PIBB algorithm learns the correlation between ProMP weights and provided feedback, thereby refining a distribution of ProMP weights utilized to generate new, improved movements.
In our study, we also developed a graphical user interface (GUI) to showcase the modalities available and to capture user interactions for our analysis.
We configured two versions of our GUI: one that incorporates only the basic modality and another that, in addition, offers all meta modalities to the participants, see Figure \ref{fig:GUIGruppe2}.

\subsection{Modalities}
\subsubsection{Base Modality Preference}
For the preference modality, two different behaviors are presented, from which the participants the one they prefer. 
We used the function 
\[ \begin{cases} 
    ( 100, 100)  & \texttt{Both preferred}\\
    ( 100,-100) & \texttt{First action preferred} \\
    (-100, 100) & \texttt{Second action preferred} \\
    (-100,-100) & \texttt{None preferred}
\end{cases}
\]
This approach not only enables the articulation of a preference for one option but also accommodates the expression of preference for neither or both options.
Our choice for this implementation was driven, among other factors, by the need for the algorithm to perform rapidly and sample-efficiently.
To amplify the differentiation in actions, we increased the disparity between positive and negative rewards, addressing the issue that users struggle to express their preferences when actions are too similar.
We also opted for samplewise updates to the sampling instead of batchwise, as users in direct interaction expect immediate responses \cite{vollmer2018user}.
\subsubsection{Meta Modalities}
The $PIBB^2$ algorithm calculates a weighted average, weighted by so called $PIBB^2$ weights.
These are produced by computing the norm from the $ProMP$ weights and the rewards they received.
We additionally incorporated a decay factor for past rewards $(=0.9)$, in addition to the covariance decay factor $(=0.973)$, which diminishes the variance of the produced distribution.
The resulting rewards are then normalized by 
\[(r_n)_i = -h * (1 - ((r_i - min(rewards)) / reward \: range))\]
with $h = 10$ the eliteness factor of $PIBB^2$.
The normalized rewards are then exponentiated which makes them $PIBB^2$.
Those are lastly normalized by the sum of weights.
\paragraph{Guidance} was achieved by significantly increasing the influence of the action chosen as guidance in the selection of future actions, while drastically diminishing the impact of all other actions.
This is achieved influencing the custom decay of rewards so it only spare the last sample marked as guidance.
In case, one of the last preference samples was marked as guidance, additionally all other rewards are decayed by a guidance decay factor of $(0.5)$. 
Furthermore, the normalization is enhanced by multiplying the eliteness factor by $1.3$.
This intensifies their influence on the normalized rewards positively due to their positive rewards. 
Lastly, they receive higher rewards of $150$ instead of $100$.

\paragraph{Correction}, as the inverse of Guidance, is achieved by significantly reducing the impact of the chosen action to deter its selection in the future. 
Unlike guidance samples, correction samples do not induce a decay, as negative actions should retain their influence in contrast to previously positive actions.
Furthermore, they are calculated differently during the normalization in the same as the Guidance since they receive negative rewards they are negatively enhanced.  
Correction samples also receive more negative rewards of $-150$ instead of $-100$.

\paragraph{Demonstrations} were done via kinesthetic teaching and served as a new average for generating subsequent actions.
    Given that users anticipate behavior closely mirroring the demonstration, past inputs are disregarded; otherwise, movements, particularly in the later stages of the learning trajectory, would deviate significantly from the demonstration due to the influence of previous actions.

\paragraph{Exploration} was facilitated by adjusting the base exploration factor upwards or downwards on five levels, ensuring, however, that exploration consistently decreases over time. 

\paragraph{Speed} was realized as accelerating or decelerating all movements after trajectories were generated, eliminating the need to incorporate an additional dimension into the learning process.

\paragraph{Fallback} was implemented by marking a movement that can be loaded by the user as needed.
When a fallback is saved it is marked as a guidance as well. 
This triggers all positive effects of a guidance since a fallback sample is marked as the best so far by the user. 
If it is loaded it has the same effect as providing a demonstration.

\subsection{Study Design}

\subsubsection{Hypotheses}
\hfill\\
\textbf{H1: Enhanced Learning through Multiple Modalities} The learning process in a human-in-the-loop reinforcement learning framework can be improved by providing multiple modalities of direct user input.\\
\textbf{H2: User Preferences and Modality Utilization} Users do not utilize all modalities equally but tend to show a preference for certain ones.\\
\textbf{H3: Differential Benefits of  Modalities} Some modalities are perceived by users to be more beneficial compared to others in facilitating the learning process.\\
\textbf{H4: Enhanced Learning through Individual Modalities} The use of individual modalities positively correlate with the efficacy of the learning algorithm when they are employed.
\subsubsection{Participants}
The participant cohort for the study was randomly assigned to two groups.
Group 1 as the baseline condition only expressed their preference between two movements and consisted of 15 individuals: 6 male, 8 female, and 1 identifying as diverse.
Group 2 was made available all additional meta-modalities and comprised 18 participants: 9 male and 9 female.
Across groups, there were 18 individuals with a higher education entrance qualification and 15 with a higher education degree, totaling 33 participants.
All 33 participants were German-speaking, aligned with the German-only interface of the system used. 
Participants were recruited on campus through flyers, posters, and mailing lists. All subjects gave their informed consent for inclusion before they participated in the study. The study was conducted in accordance with the Declaration of Helsinki, and the protocol was approved by the Ethics Committee of Bielefeld University.
\subsubsection{Task}
The task designed for the study was a minigolf challenge without obstacles, chosen for its balanced blend of enjoyment, clarity, and practicality.
The setup of the task is depicted in Figure \ref{fig:setup}.
The simplicity of the objective
made the goal and proper technique apparent for evaluation.
Furthermore, the straightforward nature of the task allowed for easy reset between attempts.
In this process, the robot's action pairs were evaluated 40 times, even if the task had been completed beforehand.
The participants were instructed to succeed (i.e., the robot hits the hole) as often as possible.
\subsubsection{Procedure}
\begin{figure}[thpb]
    \centering
    \includegraphics[scale=0.08]{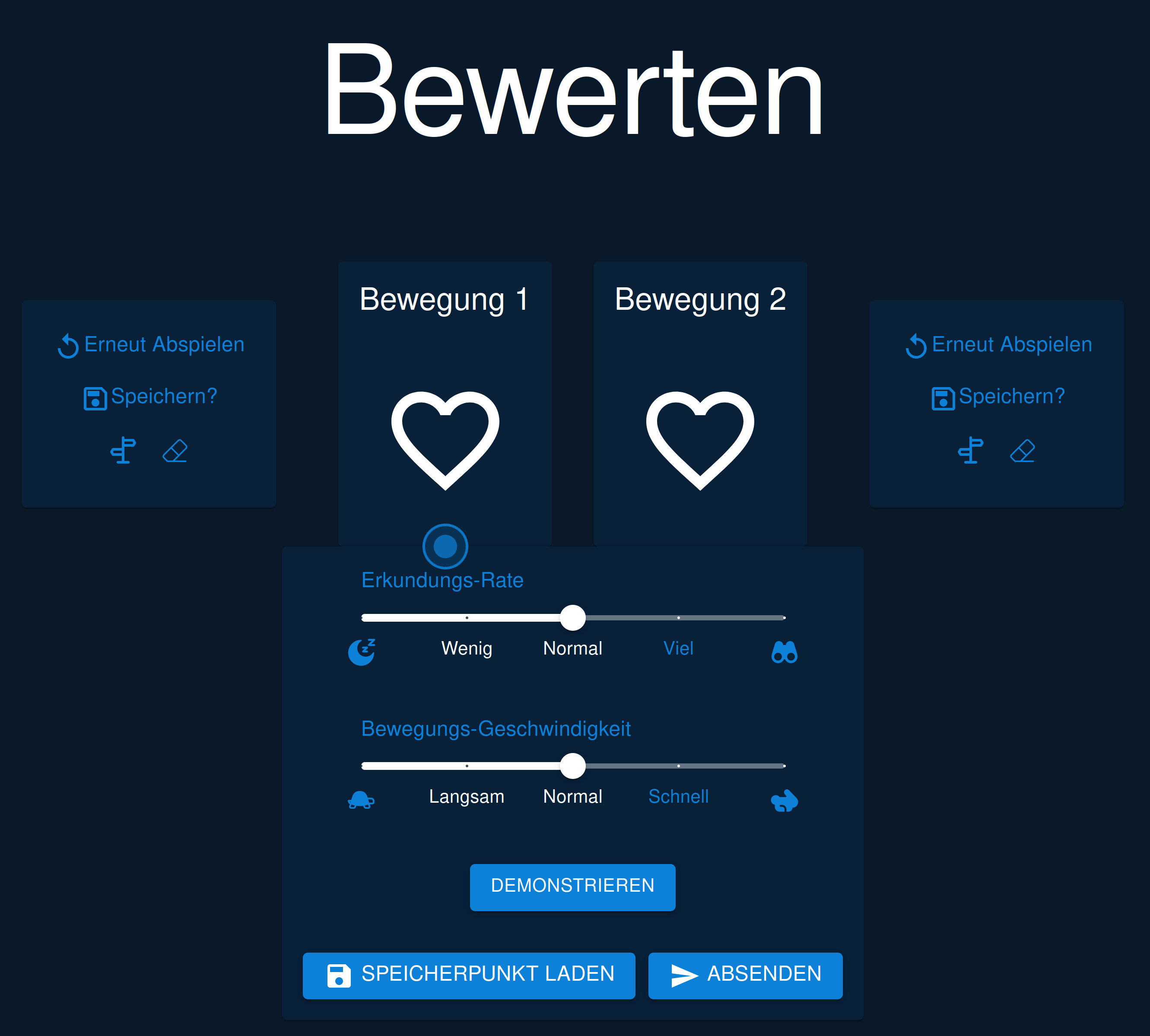}
    \caption{This Figure showcases the interface for Group 2, equipped with all meta-modalities and the base modality.}
    \label{fig:GUIGruppe2}
\end{figure}

The study began with a welcome and briefing on the task of teaching a robotic arm to play minigolf. 
The protocol detailed the use of a user interface for rating the robot's task executions.
The GUI for Group 1 was designed for simple interaction.
It displays two options, 'Bewegung 1' for the first and 'Bewegung 2' for the second movement, each with a replay button and a heart symbol for expressing preference.
At the bottom, there is a submit button for confirming choices. 
The second GUI for Group 2 was configured to evaluating robotic movements with all meta modalities, see Figure \ref{fig:GUIGruppe2}.
In addition to the features available to Group 1, it also have the option to save movements as fallback.
Furthermore, central sliders allow users to adjust exploration rates and movement speed. 
The central button at the bottom facilitates the demonstration of movements, and below on the right there is the option to load a fallback movement. 
After the introduction to the corresponding GUI, the study protocol outlined the recording of choices and timing for analysis purposes, concluding with a post-interaction questionnaire described in the following and compensation details.
\subsubsection{Data Acquisition and Analysis}
Data collection focused according to the hypotheses and study aims on two primary objectives: evaluating learning progress and assessing user satisfaction and preference.

Learning progress was analyzed through video recordings.
Our primary metric for assessing learning progress was the completion of the task, specifically measuring how often the task was completed successfully and the timing of the first success (i.e., the first hit).
We examined not only the total of successful hits compared between the two groups but also across individual trials (1-40), noting how many participants achieved a hit in each attempt.
Concerning user satisfaction, overall satisfaction with the system for both groups was measured on a 5-point Likert scale.
We calculated the significance across the groups to analyze the data.
Group 2 participants also responded to a SUS questionnaire \cite{sus} for each meta-modality.
From the calculated SUS scores, the average was determined for each meta-modality and compared pairwise. 
Preferences for the modalities were measured not only through SUS evaluations but also directly via rankings of the meta modalities.
We also collected the reasons behind these rankings.
For each option in the ranking, we calculated the percentage frequency of its placement in each position.
The qualitative statements from participants were analyzed by categorizing them into thematic groups and then evaluating these categories based on their size.
Lastly, we analyzed the relationship between learning success and modality usage.
\section{Results}


Participants in Group 2 achieved their first hit on average after 17.56 ($SD=13.65$) attempts, whereas participants in Group 1 required 34.60 ($SD=27.93$) attempts to reach their first hit.
The results from the Exact Wilcoxon-Mann-Whitney Test suggests a trend, which does not reach statistical significance of $p<0.05$ ($Z=1.7622$, $p=0.0796$). 

\begin{figure}
    \centering
    \includegraphics[scale=0.55]{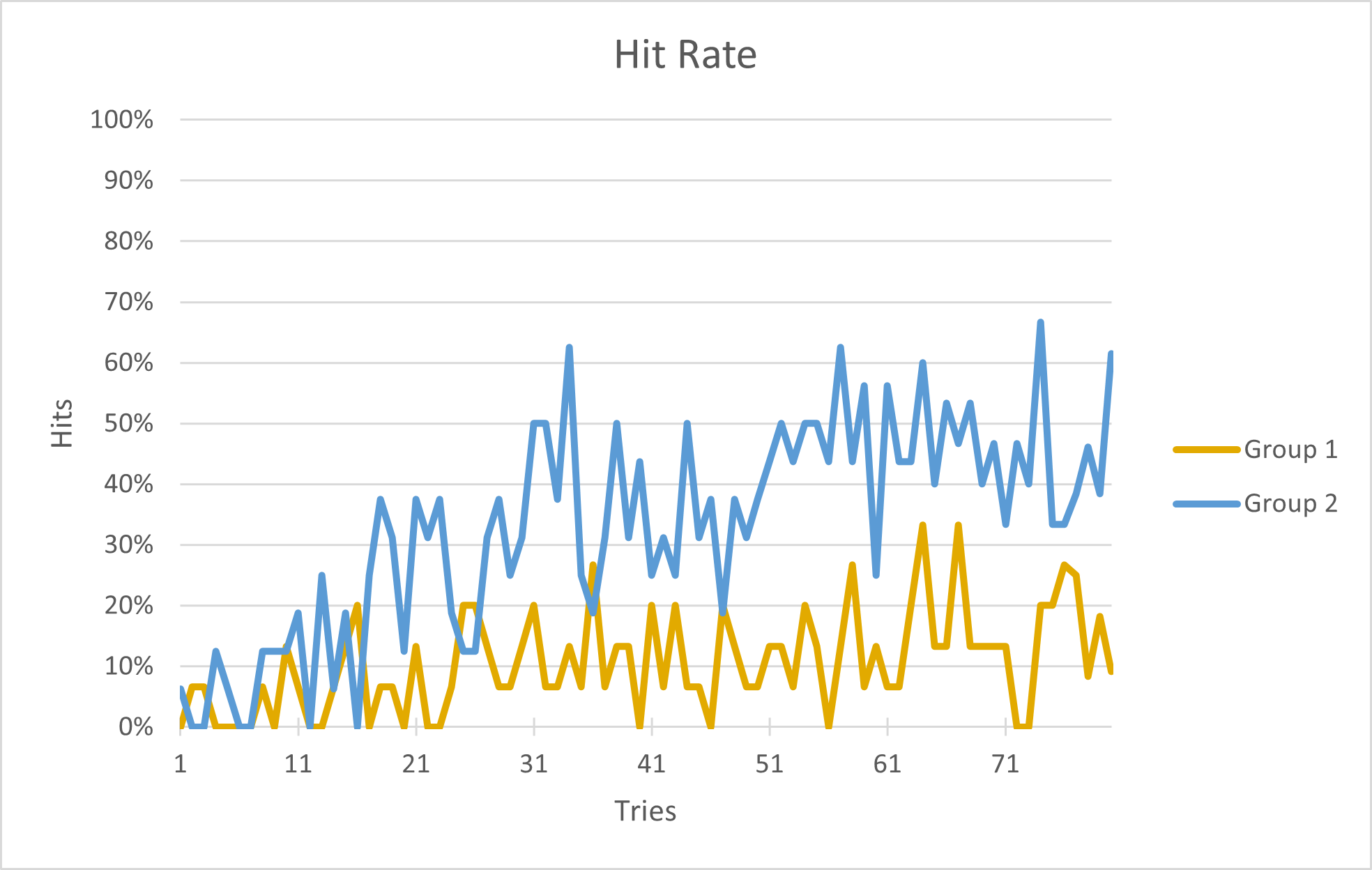}
    \caption{This graph illustrates the percentage of successful hits per trial across all participants.}
    \label{fig:Hits}
\end{figure}

Comparing the total number of hits between Group 1 and Group 2 yielded a significant difference in the performance of the two groups, with Group 2 hitting more frequent ($Z=-3.24$, $p=0.0007$) as presented in Figure \ref{fig:Hits}.




Concerning participant satisfaction, the overall satisfaction level of Group 1 was consistently moderate, with $M= 3.31$ ($SD= 1.14$).
The mean scores for overall satisfaction in Group 2 were both notably higher, standing at $M= 4.17$ ($SD= 0.857$).
The Asymptotic Kruskal-Wallis Test indicates a significant difference in satisfaction values, $\chi^2(1,34) = 5.3026$, $p <0.05$, with $p= 0.022$ , revealing that participants in Group 2 were more satisfied with the system than those in Group 1.
Due to the demonstration modality being sparsely used, we excluded it from the SUS score evaluation, as the results would not be very representative.
With mean SUS scores of $M= 79.17$ ($SD= 8.83$) for Guidance, $M= 79.67$ ($SD= 14.81$) for Speed, $M= 74.22$ ($SD= 15.80$) for Correction, $M= 72.67$ ($SD=12.12$) for Fallback, and $M= 68.82$ ($SD= 16.94$) for Exploration, the system is generally perceived as good in usability; Guidance and Speed are highly satisfactory, while Correction and Fallback are seen as good but with room for improvement, and Exploration, though still good, is identified as having the most noticeable areas for enhancement.

The Friedman test suggests a significant difference in the SUS-Score-values, with $\chi^2(4,19) = 17.777, p <0.01$ and $p = 0.0013$. 
A significant difference was identified between Exploration and Guidance ($p=0.008$), with Guidance being perceived as more usable.
Between Exploration and Speed, a significant difference was found ($p=0.0023$), with Speed being rated higher than Exploration.
A trend was observed between Fallback and Speed ($p=0.029$), suggesting that Fallback was perceived as better.
No significant differences were found in the other pairwise comparisons. 

\begin{figure}[thpb]
    \centering
    \includegraphics[scale=0.6]{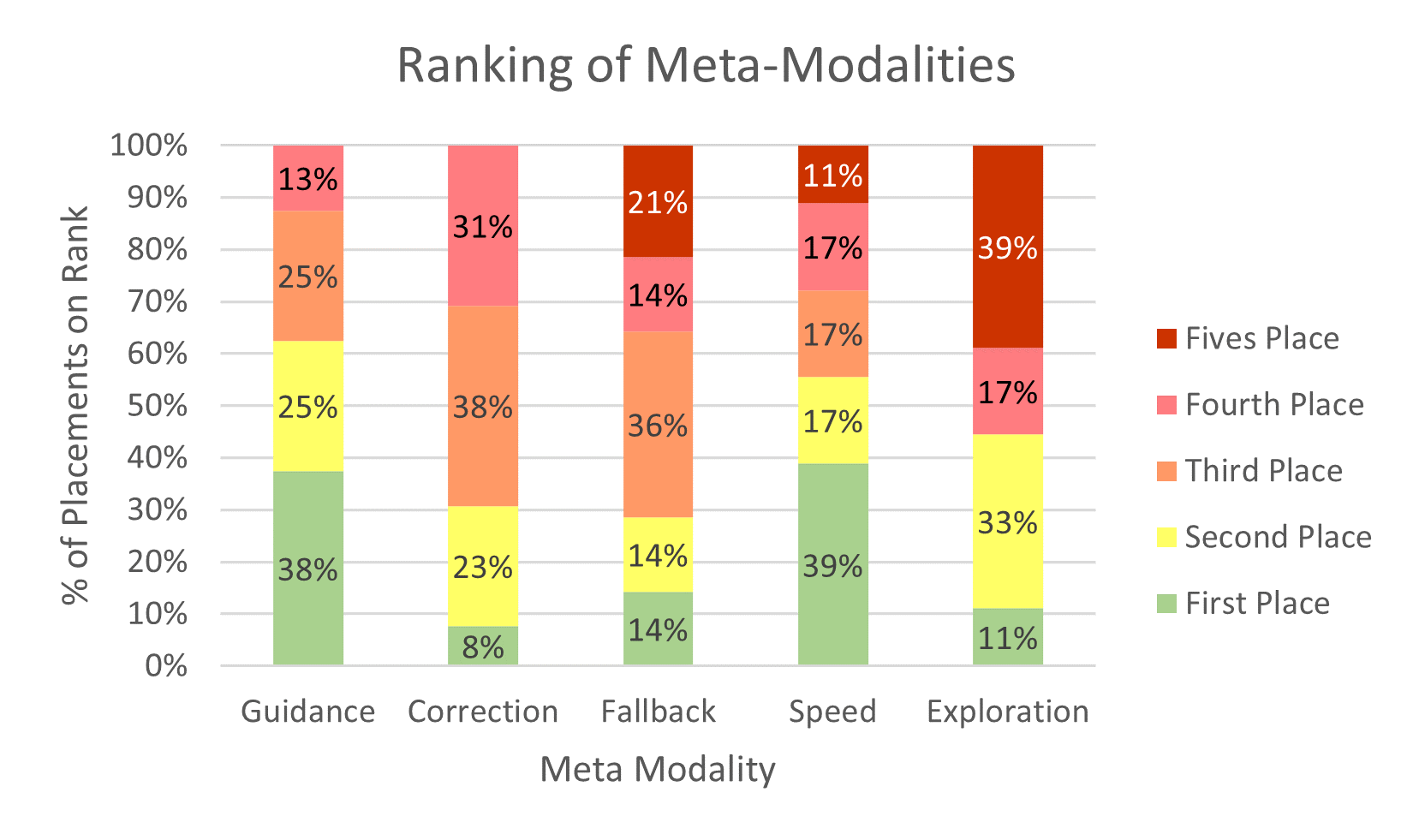}
    \caption{This diagram showcases the ranking distribution of the meta modalities.
     It illustrates the percentage placement of each meta modality across the ranking positions by Group 2 participants.}
    \label{fig:modality_ranking}
\end{figure}

We have illustrated the results of the ranking of meta-modalities in Figure \ref{fig:modality_ranking}, showing the proportion of rankings for each modality every time it was ranked.
The results indicate a preference for the ``Guidance'' modality, which was mostly placed on the first and second rank positions, showcasing its favorability among participants.
``Speed'' is favored as well, with a similar portion of its rankings in the first place, although it also shows a spread across third to even fifth ranks in contrast to ``Guidance''.
The ``Correction'' modality is ranked with moderate preference, evident from its distribution mostly across second to fourth ranks, indicating a balanced reception.
Similarly, ``Fallback'' shows a slight leaning towards the first rank but also to the fifth rank, thus its distribution suggests it is only moderately preferred.
``Exploration'' is arguably the least preferred, as the majority of its rankings are in the lower fourth and fifth ranks.
Overall, ``Guidance'' and ``Speed'' were the most liked, while ``Exploration'' is far behind in the ranking.

\begin{figure}[thpb]
    \centering
    \includegraphics[scale=0.45]{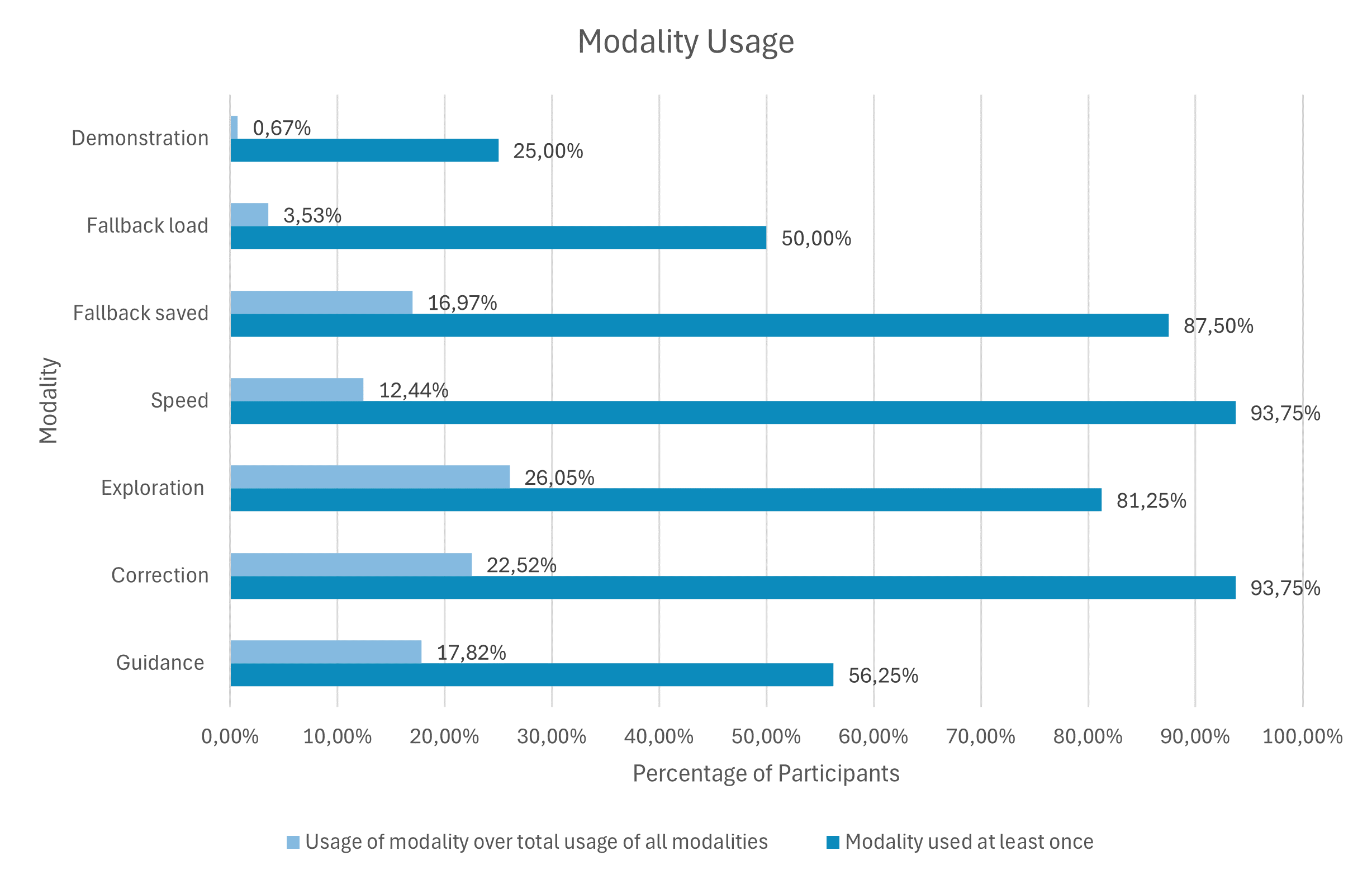}
    \caption{This Figure illustrates, on one hand, the relative usage of modalities across all clicks and, on the other, the usage among participants, taking into account whether the modality was used at least once. }
    \label{fig:Modality_Usage}
\end{figure}

Concerning the usage frequency, we evaluated two key metrics: a) the total usage means per participant where a modality, when utilized, is counted once;
b) the relative count which aggregates across all participants, measuring how frequently each modality was used overall.
The results are displayed in Figure \ref{fig:Modality_Usage}.

\begin{figure}[thpb]
    \centering
    \includegraphics[scale=0.5]{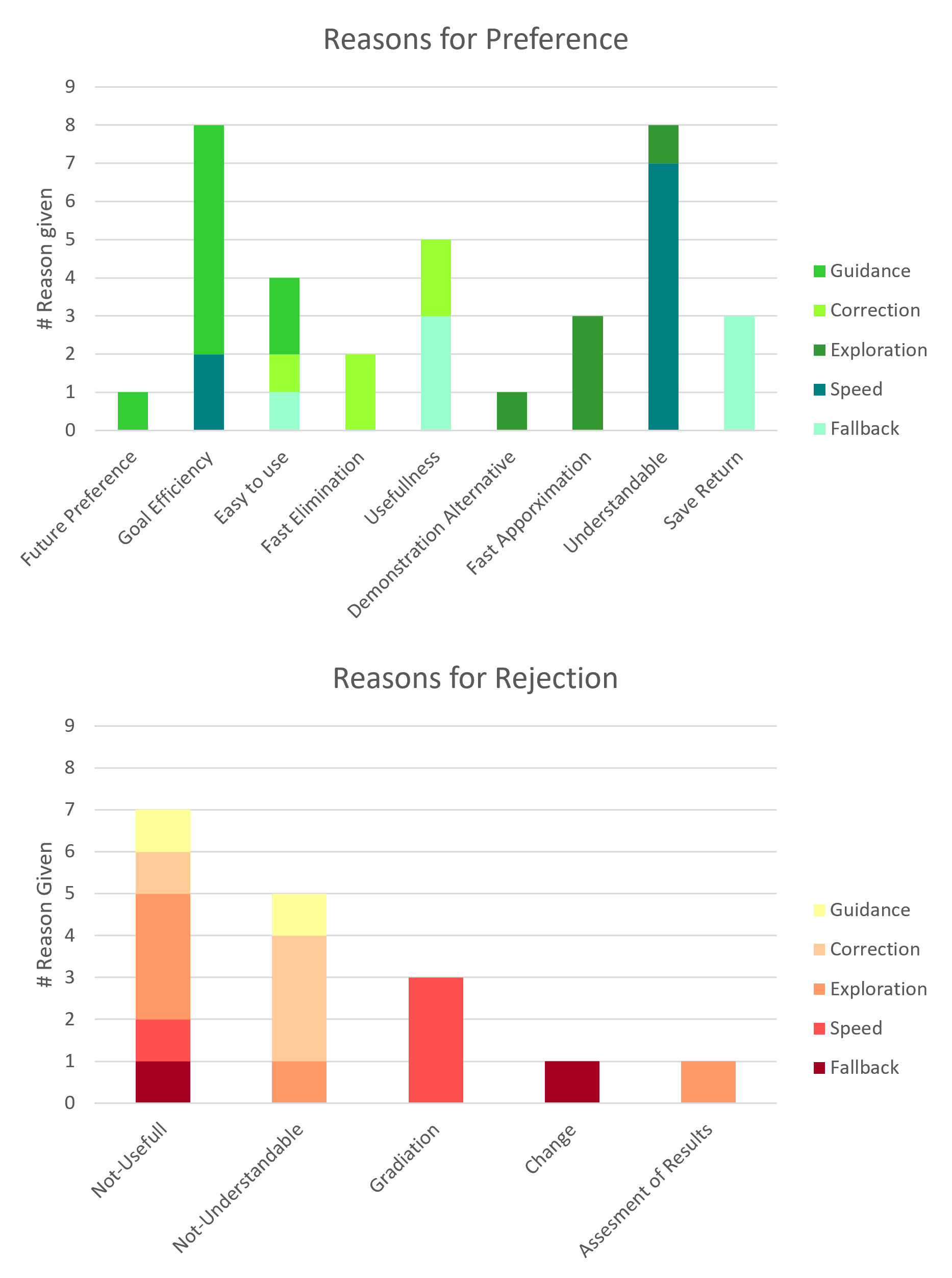}
    \caption{Presentation of Participant Rationale for Meta-Modality Rankings}
    \label{fig:reasons}
\end{figure}

The ``Demonstration'' modality saw minimal use, with just $0.67\%$ of all interactions and only $25\%$ of participants using it.
Fallback was divided into the save of a Fallback point and the load of such points.
In contrast, ``Fallback load'' was used by a higher percentage of participants ($50\%$), despite accounting for only $3.53\%$ of all interactions.
On the other hand, ``Fallback saved'' was not only used by a majority ($87.5\%$) of participants but also were a higher portion of interactions at $16.97\%$.
``Speed'' stood out with the highest engagement, utilized by $93.75\%$ of participants and making up $12.44\%$ of all interactions.
``Exploration'' was also popular, with $81.25\%$ of participants engaging with it and $26.05\%$ share of total modality use.
Similarly, ``Correction'' was used by the same percentage of participants as ``Speed'' ($93.75\%$), but with a greater interaction with $22.52\%$ participants using it.
Meanwhile, ``Guidance'' was used by $56.25\%$ of participants and $17.82\%$ of all interactions.
Overall, the ``Speed'' and ``Correction'' modalities lead in participant usage, with ``Correction'' also notable for interaction frequency alongside ``Exploration'', whereas ``Demonstration'' lags behind with the lowest interaction rate and participant engagement.\\
As previously described, we also captured the participants' explanations for the rankings analyzed above.
These are depicted in Figure \ref{fig:reasons}, categorized into positive justifications for higher placements and more critical reasons for lower rankings.
Overall, the reason that was most often given for higher rankings was ``Goal Efficiency'' and ``Understandability''. Due to space restrictions, for all other rankings, we refer the reader to the figure.
Spearman's rank correlation was used to analyze the correlations of hits per try and modalities used per try.
A correlation between Exploration and hit success unveiled a statistically significant, albeit weak, negative correlation, with $(\rho =-0.21, p = 0.025)$. 
Thus, with a decrease of Exploration
the likelihood of hitting the target marginally increases.
The analysis of Speed revealed a modest positive correlation $(\rho = 0.2389221, p = 0.03898)$. 
This suggests that higher speeds correlate with a slightly improved chance of hitting the target.
Contrastingly, the correlation with load $(\rho = -0.01)$ and save $(\rho = 0.007)$ of a fallback were not only very weak, but both also lacked statistical significance with $(p = 0.7)$ and $(p = 0.8)$, indicating an essentially negligible relationship between fallback and hit success.
Guidance also showed a weak negative correlation with hit success, with ($\rho =  -0.03$), the finding was as well not statistically significant $(p = 0.3)$, suggesting that guidance does not significantly influence hit success.
Between Correction and hit success a statistically significant negative correlation ($\rho = -0.07$, $ p = 0.01$) was found.
However, the correlation is below $0.1$ thus can not even be considered as weak. 

Overall, the analyses show that while an increase in speed and a decrease in exploration positively influences hit success fallback, guidance and correction do not show significant or notable correlations with hit success.

\section{Discussion}
\textbf{H1: Enhanced Learning through Multiple Modalities.} Our results affirm that providing multiple feedback modalities significantly enhances the learning process, as evidenced by Group 2's higher performance. 
This group consistently hitting targets confirms that multiple modalities not only expedite initial learning success but also fortify sustained learning progress.

\textbf{H2: User Preferences and Modality Utilization.} Clearly users do not engage with all modalities equally, displaying marked preferences.
These preferences are mirrored in the higher overall satisfaction observed in Group 2, suggesting that the success facilitated by preferred modalities positively influences user perceptions.
Participants expressed their preferences through usage frequency, rankings, reasons for their choices, and SUS scores of the individual meta-modalities.
Notably, Guidance and Speed stand out for their high usage and positive reception, attributed to their perceived efficacy impact on the learning process and intuitiveness.
Both also stand out in the pairwise comparison regarding the SUS score compared to Exploration.
Conversely, Exploration and Demonstration were less favored, with the former criticized for its ambiguous effects and the latter for its underuse, possibly due to perceived effort or lack of necessity.

\textbf{H3: Differential Benefits of Modalities.} Our findings corroborate the hypothesis that certain modalities are perceived as more beneficial than others in enhancing the learning process.
Guidance, with its high engagement and favorable ranking, was seen as particularly effective, emphasizing its role in achieving learning goals efficiently.
Although Correction and Fallback were recognized for their utility, their engagement levels and mixed rankings suggest a nuanced perception of their effectiveness in learning enhancement.

\textbf{H4: Enhanced Learning through Individual Modalities} Exploration and Speed positively impact hit success, with exploration reduced upon successful hits to maintain performance and speed ensuring the ball's reach. 
Guidance emerged as the most valued by users, while Exploration was seen as least useful.
However, if its function was understood, it is enhancing learning success.
Movement attributes, perceived positively and frequently interacted with, and thus seem to be a natural teaching method for participants even though they are underestimated in the field. 

In summary, our findings advocate for a broader adoption and integration of multiple modalities within the domain of interactive robot learning.
Particular focus should be placed on clarifying exploration, since it positively correlates with learning progress when used correctly, on leveraging intuitive movement properties, and on emphasizing guidance, which is a favorite among participants.

\addtolength{\textheight}{-12cm}   




\bibliographystyle{ieeetr}
\bibliography{biblio}
\end{document}